\documentclass{article} 
\pdfoutput=1
\usepackage{lmodern} 
\usepackage{iclr2027_conference,times}

\iclrfinalcopy

\usepackage{amsmath,amsfonts,bm}

\def\eqref#1{equation~\ref{#1}}

\def\1{\bm{1}}

\DeclareMathAlphabet{\mathsfit}{\encodingdefault}{\sfdefault}{m}{sl}
\SetMathAlphabet{\mathsfit}{bold}{\encodingdefault}{\sfdefault}{bx}{n}

\usepackage{booktabs}
\usepackage{tabularx}
\usepackage{xcolor}
\usepackage{colortbl}
\definecolor{tablemint}{HTML}{E1F1EC}
\definecolor{tablelavender}{HTML}{E3DEEE}
\definecolor{tableblue}{HTML}{DCEAF7}
\definecolor{tablecream}{HTML}{FEECDD}
\definecolor{tableink}{HTML}{344454}
\newcommand{\tablesr}[2]{\shortstack[c]{$#1$\\[1pt]{\scriptsize$\pm #2$}}}
\usepackage{float}
\usepackage{placeins}
\usepackage{hyperref}
\usepackage{url}
\usepackage{microtype}
\usepackage{graphicx}
\usepackage{wrapfig}
\usepackage{amssymb}

\title{Beyond Actions: Learning Future Operation Targets for Vision-Language-Action Models}

\author{
\parbox{\textwidth}{%
\centering
\vspace*{10pt}
\textbf{Min Lin$^{1}$\thanks{Equal contribution.},
Bokai Zhou$^{1}$\footnotemark[1],
Jianqi Lin$^{1}$,
Kaiyuan Yang$^{1}$}\\[3pt]
\textbf{Yicheng Chang$^{1}$,
Junjie Yu$^{1}$,
Xiaofan Liu$^{1}$,
Jianhua Han$^{2}$}\\[3pt]
\textbf{Rui Li$^{2}$,
Shen Zhao$^{1}$%
\thanks{Corresponding authors: Shen Zhao and Xiaodan Liang.},
Xiaodan Liang$^{1,2}$\footnotemark[2]}\\[8pt]
{\normalfont
$^{1}$Sun Yat-sen University\\[2pt]
$^{2}$Yinwang Intelligent Technology Co., Ltd.\\[3pt]
\texttt{linm57@mail2.sysu.edu.cn}
}
}
}
\begin{document}

\maketitle
\fancyhead{}
\renewcommand{\headrulewidth}{0pt}

\begin{abstract}
Actions specify what a robot should execute, but do not describe what those commands will achieve. Vision-language-action (VLA) policies nevertheless learn primarily by predicting action chunks from images, language, and state. We introduce Action--Future Coupled Effect Learning (AFCE), which makes a future-grounded physical Effect the policy's primary prediction target. An offline tokenizer constructs temporally ordered Effect tokens from demonstrated actions, realized robot responses, and sparse visual transitions. Reconstruction objectives encourage the representation to retain executable detail and observed physical change. A flow-matching VLA predicts Effects from current images, instruction, and robot state, while a state-conditioned decoder converts them into action chunks. Embodiment-specific interfaces allow single- and dual-arm tasks to share the Effect model. On DexJoCo, AFCE achieves $50.97\%$ overall success, improving over $\pi_{0.5}$ and UVT by 11.82 and 15.21 percentage points, respectively. It leads on all five bimanual tasks, highlighting the value of Effect supervision for coordinated dexterous manipulation. The project page is available at \url{https://afceproject.github.io/AFCE-Web/}.
\end{abstract}

\section{Introduction}
\label{sec:introduction}

Vision-language-action (VLA) models have made substantial progress toward generalist robot control by mapping images, language, and robot state to continuous actions \citep{zitkovich2023rt2,octo2024,kim2025openvla,black2024pi0,black2025pi05}. Yet the target used to train this mapping usually describes \emph{what the robot commanded}, not \emph{what the command changed}. An action sequence specifies motor behavior, but leaves its effects on the robot and environment implicit. In dexterous manipulation, this gap matters: contact, hand configuration, and object pose can turn similar commands into different outcomes, while different motions can achieve the same task-relevant change. The challenge is to make physical progress part of the policy's prediction target without sacrificing the detail needed to execute an action.

Predicting the future offers one way to expose such progress. World and world-action models use future observations or latent dynamics to support control \citep{wu2024gr1,cen2025worldvla,zhang2025dreamvla,ye2026dreamzero}. However, a complete visual future contains changes due to appearance, occlusion, and camera motion that need not determine the next control command. Conversely, a scene change alone may leave out the contact timing, finger coordination, and robot motion required to produce it. Latent-action and future-informed visuomotor methods provide more compact targets \citep{ye2024lapa,bu2025univla,chen2025villax,luo2026jala, liu2026lara,xu2026futurevla,luo2026beingh07}, but compactness alone does not resolve whether a target is both predictable from the current observation and sufficient for precise action decoding.

We therefore seek a target that connects three requirements: it must be predictable before acting, retain task-relevant information about the resulting physical transition, and support recovery of executable continuous actions. We introduce \textbf{AFCE (Action--Future Coupled Effect Learning)}, which learns a temporal \emph{Effect} from synchronized demonstrations of executed actions, realized robot-state changes, and sparse world changes. Rather than specifying the Effect through hand-designed semantic labels, we let these sources interact and constrain a shared representation through joint reconstruction. The action objective retains control detail, while the robot-response and world-change objectives ground that detail in observed consequences.

\begin{figure}[t]
    \centering
    \includegraphics[width=1.0\textwidth]{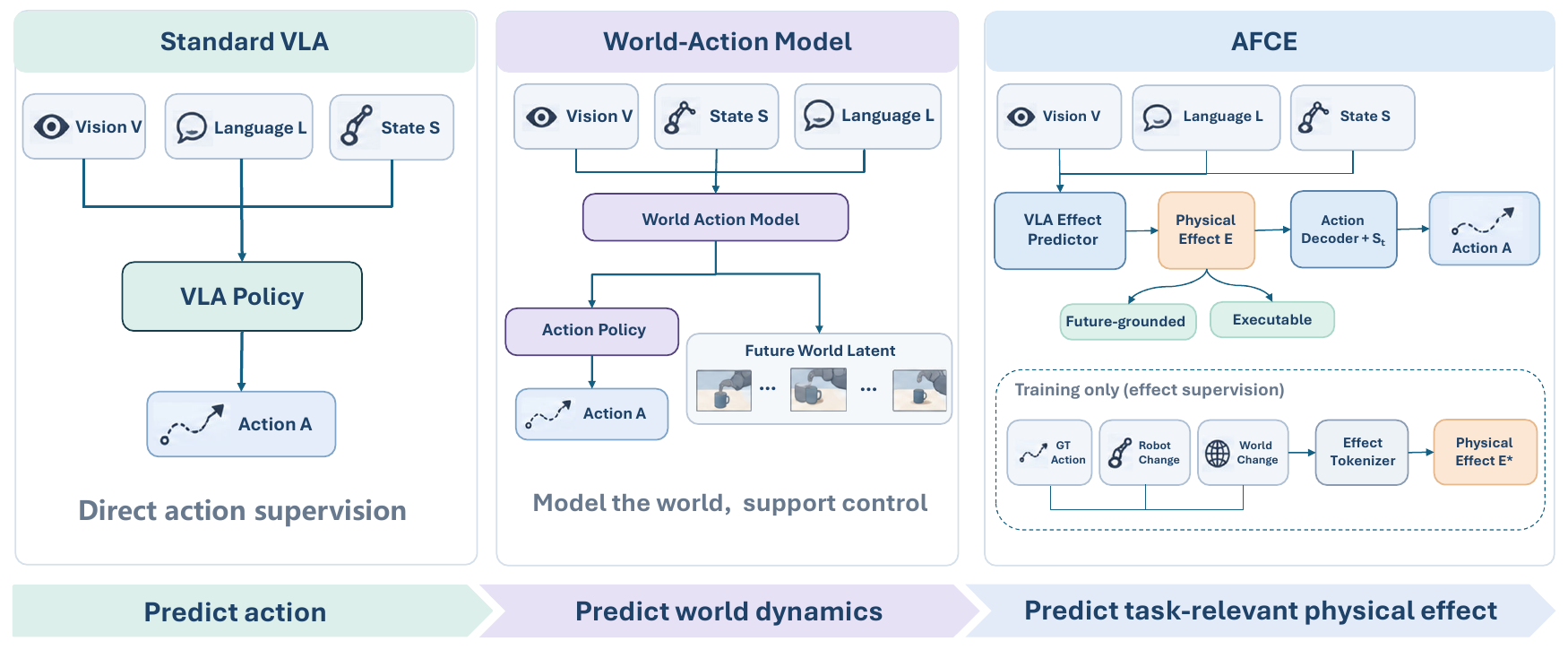}
    \caption{
    Comparison of policy supervision paradigms. A standard VLA
    predicts actions directly; world-action models use future
    dynamics to support control; AFCE predicts an action-decodable
    Effect grounded in executed actions, robot response, and world
    change. Future evidence constructs training targets but is
    unavailable to the deployed policy.
    }
    \vspace{-15pt}
    \label{fig:framework}
\end{figure}

Figure~\ref{fig:framework} illustrates the resulting training interface. An offline tokenizer constructs Effect targets from recorded trajectories, where future evidence is available. The VLA then learns to predict these targets using only the current images, language instruction, and robot state. At deployment, a decoder maps the predicted Effect and current state to an action chunk. Thus, future observations shape the supervision without becoming inputs to the deployed policy.

The central test of this interface is whether the Effect remains useful \emph{after prediction}. Accurate reconstruction from a ground-truth Effect does not ensure that the VLA can predict that Effect, nor that decoded actions succeed under closed-loop control. We evaluate these questions separately: reconstruction tests what the representation retains; ground-truth-Effect replay tests whether it supports execution; and predicted-Effect rollout tests the complete policy against direct action supervision under the same evaluation protocol. In mixed single- and dual-arm training, AFCE improves aggregate task success, particularly on dual-arm tasks, while some fine-control tasks remain challenging.

Our contributions are:
\begin{itemize}
    \item We formulate VLA target learning around three coupled requirements: predictability from current observations, information about task-relevant future changes, and continuous-action executability.

    \item We introduce AFCE, a temporal Effect learned by jointly reconstructing executed actions, realized robot response, and sparse world change from a shared representation.

    \item We evaluate the representation and the complete policy through reconstruction, ground-truth-Effect replay, and predicted-Effect closed-loop control, showing aggregate gains under mixed single- and dual-arm training while identifying remaining fine-control limitations.
\end{itemize}

\section{Related Work}
\label{sec:related-work}

\subsection{VLA Policies and Action Supervision}

Vision-language-action (VLA) models have established a powerful interface between semantic understanding and robot control. RT-2 casts discretized actions as language-like tokens, while Octo and OpenVLA learn generalist policies from heterogeneous robot data \citep{zitkovich2023rt2,octo2024,kim2025openvla}. $\pi_0$ and $\pi_{0.5}$ further combine pretrained visual-language representations with a flow-matching action expert for continuous control \citep{black2024pi0,black2025pi05}. Complementary advances such as ACT, Diffusion Policy, and FAST improve action prediction through chunking, expressive generative modeling, and efficient tokenization \citep{zhao2023act,chi2023diffusion,pertsch2025fast}. Together, these methods provide increasingly capable ways to generate robot commands. They also motivate a closer examination of what the policy is asked to predict: an action records the commanded motion, whereas the resulting robot and scene changes carry additional information about the intended manipulation. Building on these advances, AFCE preserves the VLA observation interface and continuous action execution while introducing a physical Effect as the policy's intermediate supervision target.

\subsection{World Models and Future-Informed Control}

Future-aware learning offers a complementary route to stronger visuomotor policies. GR-1 and VidMan use video dynamics to inform manipulation, while WorldVLA and DreamZero connect predictions about the evolving world to action generation \citep{wu2024gr1,wen2024vidman,cen2025worldvla,ye2026dreamzero}. FutureVLA and Being-H0.7 further show how future-informed representations can guide policies that act from current observations \citep{xu2026futurevla,luo2026beingh07}. Collectively, these studies establish future information as a valuable training signal for control. AFCE asks how that signal can be organized into a target that is both informative about physical change and directly useful for producing actions. Demonstrated actions, realized robot responses, and sparse visual transitions jointly specify the Effect during training; at deployment, the policy predicts it from the current context and a state-conditioned decoder converts it into robot commands. This formulation explicitly links future-grounded supervision to the action interface used for closed-loop execution.

\subsection{Latent Actions and Transferable Policy Targets}

Latent-action learning provides intermediate targets that capture structure beyond individual action coordinates. LAPA derives latent actions from visual transitions, while UniVLA and villa-X develop representations for learning from robot and video data \citep{ye2024lapa,bu2025univla,chen2025villax}. JALA and LARA connect transition-derived features to actions and policy learning \citep{luo2026jala,liu2026lara}. UVT combines demonstrated action chunks with discrete visual dynamics codes into a decodable policy target \citep{feng2026unifiedvisuomotortargetssupervising}. AFCE develops this direction by constructing a temporal Effect from demonstrated commands, realized robot responses, and sparse visual transitions. Its action and response branches relate observed scene changes to both the issued commands and the motion actually realized. Realized responses are particularly informative in dexterous manipulation, where contact and hand configuration can make similar commands produce different outcomes. Temporal Effect tokens preserve the progression from initial contact to subsequent object interaction. A state-conditioned decoder maps the Effect to native actions, making the learned target useful for continuous control. Embodiment-specific adapters accommodate different native state and action dimensions while retaining a common Effect shape.

\section{Action--Future Coupled Effect Learning}
\label{sec:method}

\begin{figure}[t]
    \centering
    \includegraphics[width=\linewidth]{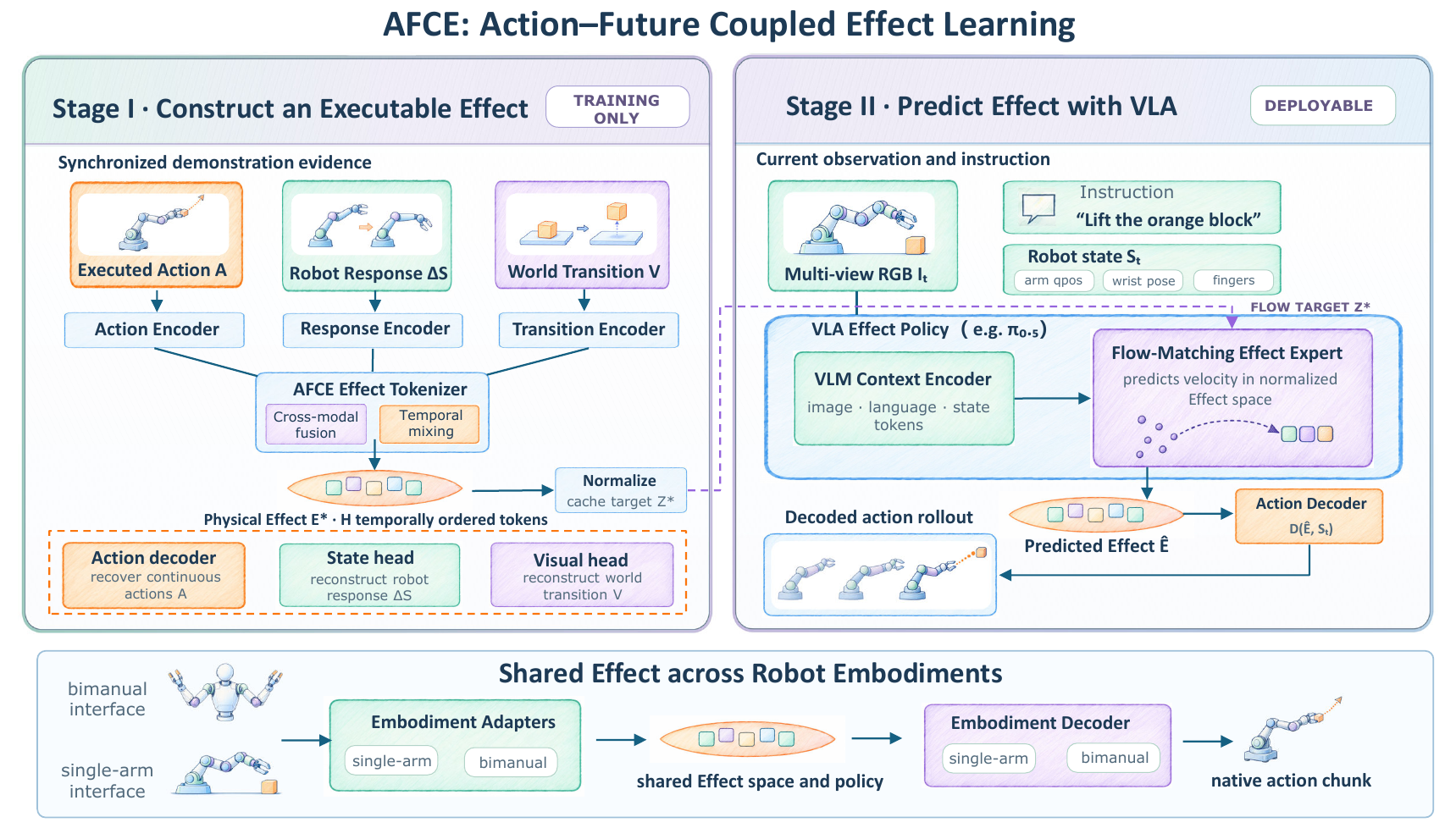}
    \caption{\textbf{Overview of Action--Future Coupled Effect Learning (AFCE).}
    Stage~I fuses demonstrated actions $\mathbf{A}_t$, robot response
    $\Delta\mathbf{S}_t$, and visual transitions $\mathbf{V}_t$
    into ordered Effect tokens $\mathbf{E}_t^{*}$, supervised by
    action, state, and visual reconstruction. Stage~II predicts the
    normalized target $\mathbf{Z}_t^{*}$ from current images,
    instruction, and robot state, then decodes an action chunk.
    The lower panel shows the shared Effect space with
    embodiment-specific interfaces. The dashed path denotes
    training targets; future evidence is not used at deployment.}
    \label{fig:afce-overview}
    \vspace{-15pt}
\end{figure}

We consider language-conditioned manipulation with current observation $\mathbf{o}_t=(\mathbf{I}_t,\ell,\mathbf{s}_t)$, where $\mathbf{I}_t$ denotes the images, $\ell$ the instruction, and $\mathbf{s}_t$ the robot state. A conventional VLA predicts an action chunk $\mathbf{A}_t=[\mathbf{a}_t,\ldots,\mathbf{a}_{t+H-1}]$ directly from $\mathbf{o}_t$. AFCE instead trains the policy to predict a temporal \emph{Effect}: a compact target constructed from issued actions, the robot's realized response, and changes in the observed world. The complete framework is shown in Figure~\ref{fig:afce-overview}.

For each demonstration window, let $\Delta\mathbf{S}_t$ denote the realized robot-response sequence and $\mathbf{V}_t$ sparse visual transition evidence over $[t,t+H]$. Stage~I constructs
\begin{equation}
    \mathbf{E}_t^{*}
    =T_{\phi}\!\left(
        \mathbf{A}_t,\Delta\mathbf{S}_t,\mathbf{V}_t
      \right)
    \in\mathbb{R}^{H\times D_E},
    \label{eq:effect-definition}
\end{equation}
where $T_{\phi}$ is the Effect tokenizer. We use $H=30$ and $D_E=256$, yielding one ordered token per action step. The superscript ${*}$ distinguishes a target obtained from demonstration evidence from an Effect predicted at deployment.

Stage~I learns the tokenizer with action and transition reconstruction. We then freeze it, cache its Effect outputs, and use fixed statistics to obtain normalized targets $\mathbf{Z}_t^{*}$ for Stage~II. The VLA predicts these targets from $\mathbf{o}_t$ while the action decoder adapts to predicted Effects. At deployment, the policy samples a normalized Effect and decodes its denormalized counterpart into actions:
\begin{equation}
    \widehat{\mathbf{Z}}_t
    \sim p_{\theta}(\mathbf{Z}\mid\mathbf{o}_t),
    \qquad
    \widehat{\mathbf{A}}_t
    =D_{\psi}\!\left(
        \mathcal{N}^{-1}(\widehat{\mathbf{Z}}_t),
        \mathbf{s}_t
      \right).
    \label{eq:afce-policy}
\end{equation}
Here $p_{\theta}$ is implemented by a conditional flow-matching VLA and $\mathcal{N}$ is the fixed Effect normalization. Future observations enter Stage~I target construction only; the deployed path uses current images, instruction, and robot state.

\subsection{Stage I: Learning an Executable, Future-Grounded Effect}
\label{sec:stage1-effect}

\paragraph{Synchronized evidence.} The three evidence sources describe distinct aspects of the same interaction. The action encoder uses group-specific projections for position, rotation, and finger commands before combining them into step-level features. The response encoder represents realized palm translation, relative orientation, fingertip displacement, and finger-joint change in the reference frame of the prediction window. Its features are computed from observed states and forward kinematics and aligned with future action steps; both arms are retained for dual-arm demonstrations. Separating commands from realized motion allows the Effect to describe the consequences of execution rather than merely re-encode the commands.

For world transitions, we sample frozen DINOv3 features at sparse times. A localized visual token records appearance-feature change, image-plane displacement, spatial location, and temporal extent. The evidence may contain robot as well as object motion and does not require perfect object segmentation. Temporal metadata associates a transition with the action steps covered by its observation interval; a change spanning several steps is not treated as an independent observation at every control instant. The exact feature construction and temporal assignment are given in the appendix.

\paragraph{Fusion before compression.} The encoders map their inputs to a common width, producing action and response features $\mathbf{z}^{A}_k$ and $\mathbf{z}^{S}_k$, together with visual features $\mathcal{Z}^{V}_k$ associated with step $k$. Modality and temporal embeddings identify their sources and positions. A learned step query first fuses the available evidence through cross-attention; a temporal Transformer then exchanges information across the full window:
\begin{align}
    \mathbf{u}_k
    &=F_{\mathrm{fuse}}\!\left(
        \mathbf{q}_k;
        \mathbf{z}^{A}_k,\mathbf{z}^{S}_k,
        \mathcal{Z}^{V}_k
      \right), \nonumber\\
    \mathbf{E}_t^{*}
    &=F_{\mathrm{temp}}\!\left(
        [\mathbf{u}_0,\ldots,\mathbf{u}_{H-1}]
      \right).
    \label{eq:effect-fusion}
\end{align}
Thus, action, response, and world-transition evidence interact \emph{before} the final Effect tokens are formed. Each token keeps its temporal position while incorporating context from other steps. Because $T_{\phi}$ processes complete training windows, its temporal attention need not be causal; this does not expose future evidence to the deployed policy.

\paragraph{Executable-bottleneck supervision.} A state-conditioned decoder reconstructs the demonstrated actions:
\begin{equation}
    \widetilde{\mathbf{A}}_t
    =D_{\psi}(\mathbf{E}_t^{*},\mathbf{s}_t).
    \label{eq:stage1-action-decoding}
\end{equation}
Auxiliary state and visual heads reconstruct the realized robot response and sparse visual transitions. The visual head receives the spatial and temporal query metadata specifying \emph{where and when} to decode, but not the target feature change. We minimize
\begin{equation}
    \mathcal{L}_{\mathrm{I}}
    =\mathcal{L}_{A}
     +\lambda_S(u)\mathcal{L}_{S}
     +\lambda_V(u)\mathcal{L}_{V}
     +\lambda_R\mathcal{L}_{\mathrm{rot}},
    \label{eq:stage1-loss}
\end{equation}
with the grouped \emph{Huber} action-reconstruction loss
\begin{equation}
    \mathcal{L}_{A}
    =\frac{
        2\mathcal{H}_{p}
        +\mathcal{H}_{r}
        +\mathcal{H}_{f}
      }{3}.
    \label{eq:grouped-action-loss}
\end{equation}
Here $p$, $r$, and $f$ denote position, rotation, and finger groups, and $\mathcal{H}_{g}$ is the corresponding normalized Huber error. Each group is averaged over valid coordinates and time steps so its dimensionality does not set its effective weight. The denominator remains three despite the position weight of two. The state and visual losses likewise average errors over valid, normalized targets. $\mathcal{L}_{\mathrm{rot}}$ is an SO(3) chordal loss that regularizes decoded orientations in addition to rotation-coordinate supervision. Action reconstruction preserves executable detail, whereas state and visual reconstruction encourage the Effect to retain realized physical changes.

\begin{wrapfigure}[17]{r}{0.55\textwidth}
    \centering
    \vspace{-15pt}
    \includegraphics[width=\linewidth]{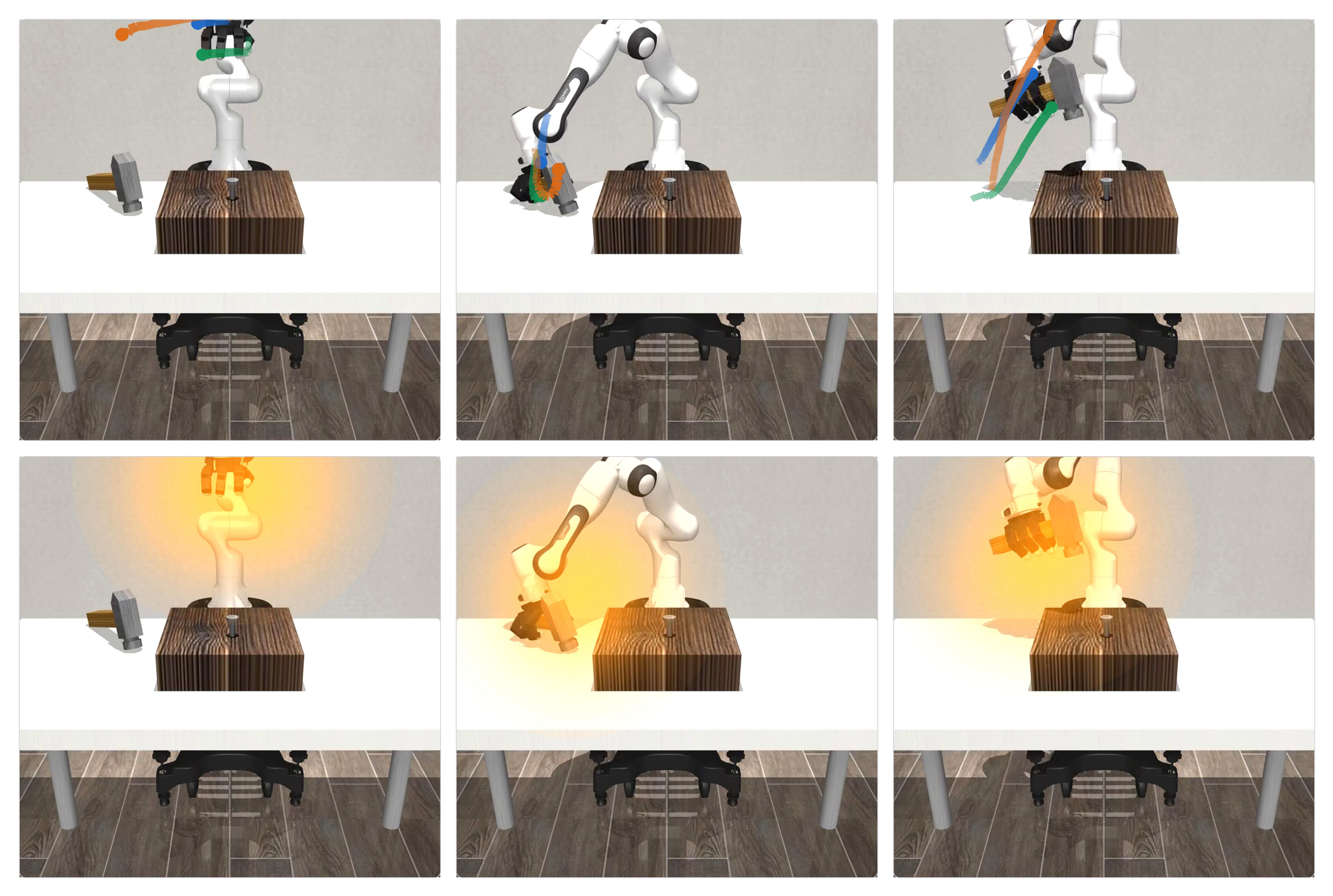}
    \vspace{-20pt}
    \caption{\textbf{Action and scene changes represented by the Effect.}
    Three successive hammer-nail frames. Top: decoded robot motion;
    bottom: accompanying scene changes.}
    \label{fig:effect-decode}
    \vspace{-8pt}
\end{wrapfigure}
Figure~\ref{fig:effect-decode} shows a hammer--nail example in which decoded robot motion accompanies visual scene change. The decoder reads the Effect and current robot state, rather than the future frame, to produce the illustrated motion. This example makes the link between the learned representation and executable action concrete, while the accompanying scene images show the physical transition used to construct its training target. The visualization is qualitative: successful reconstruction here does not establish that every evidence source contributes to the Effect, or that the policy can predict it accurately. We examine those questions through modality interventions, GT-Effect replay, and closed-loop execution.

\paragraph{Fixed targets for Stage II.} After Stage~I, we freeze $T_{\phi}$ and estimate channel-wise mean $\boldsymbol{\mu}_E$ and standard deviation $\boldsymbol{\sigma}_E$ on all Stage-I training demonstrations. The normalized target is
\begin{equation}
    \mathbf{Z}_t^{*}
    =\mathcal{N}(\mathbf{E}_t^{*})
    =\frac{\mathbf{E}_t^{*}-\boldsymbol{\mu}_E}
           {\max(\boldsymbol{\sigma}_E,\sigma_{\min})},
    \label{eq:effect-normalization}
\end{equation}
where the maximum is elementwise and $\sigma_{\min}$ is a fixed standard-deviation floor. Statistics are shared across temporal positions and remain fixed throughout Stage~II. We cache the Stage-I Effect outputs for policy training and apply this fixed normalization to form targets. The future-evidence encoders and auxiliary heads are not used at deployment.

\subsection{Stage II: Predicting and Executing Effects}
\label{sec:stage2-policy}

\paragraph{Conditional Effect flow matching.} We adapt $\pi_{0.5}$ to predict normalized Effects: its action input and output projections are replaced with projections of width $D_E$, while its visual-language-state conditioning and flow-matching backbone are retained. Let $\boldsymbol{\epsilon}\sim\mathcal{N}(\mathbf{0},\mathbf{I})$ have the shape of $\mathbf{Z}_t^{*}$ and let $\tau\in[0,1]$ be flow time, distinct from environment time $t$. We interpolate from the clean target at $\tau=0$ to noise at $\tau=1$:
\begin{equation}
    \mathbf{Z}_{t,\tau}
    =(1-\tau)\mathbf{Z}_t^{*}
      +\tau\boldsymbol{\epsilon},
    \qquad
    \mathbf{v}^{*}
    =\boldsymbol{\epsilon}-\mathbf{Z}_t^{*}.
    \label{eq:flow-interpolation}
\end{equation}
The conditional velocity field is trained with
\begin{equation}
    \mathcal{L}_{\mathrm{FM}}
    =\mathbb{E}_{t,\tau,\boldsymbol{\epsilon}}
      \left[
        \frac{1}{HD_E}
        \left\|
          \mathbf{v}_{\theta}
            (\mathbf{Z}_{t,\tau},\tau;\mathbf{o}_t)
          -\mathbf{v}^{*}
        \right\|_F^2
      \right].
    \label{eq:effect-flow}
\end{equation}
At inference, integrating the learned flow from noise toward $\tau=0$ gives a normalized predicted Effect.

\paragraph{Action-grounded prediction.} Distance in Effect space need not reflect the action error caused by decoding. We therefore supervise a clean-endpoint estimate and its decoded action chunk:
\begin{align}
    \widehat{\mathbf{Z}}_{t,0}^{(\tau)}
    &=\mathbf{Z}_{t,\tau}
      -\tau\mathbf{v}_{\theta}
        (\mathbf{Z}_{t,\tau},\tau;\mathbf{o}_t), \nonumber\\
    \widehat{\mathbf{A}}_t^{(\tau)}
    &=D_{\psi}\!\left(
        \mathcal{N}^{-1}
          (\widehat{\mathbf{Z}}_{t,0}^{(\tau)}),
        \mathbf{s}_t
      \right).
    \label{eq:clean-effect-decoding}
\end{align}
This estimate follows from Equation~\ref{eq:flow-interpolation} and avoids differentiating through a complete sampling trajectory. The ground-truth-action auxiliary loss $\mathcal{L}_{\mathrm{GT}}$ compares $\widehat{\mathbf{A}}_t^{(\tau)}$ with the \emph{demonstrated} action after fixed action standardization. It averages position, rotation, and finger Huber errors with equal group weights. Unlike Equation~\ref{eq:grouped-action-loss}, it does not apply the Stage-I position multiplier or SO(3) term.

The decoder is initialized from Stage~I and updated jointly with the policy at a reduced learning rate. Thus, it adapts to predicted Effects, and action-loss gradients also reach the policy. The same decoder is used during supervision and deployment.

We additionally match changes in finger commands across consecutive steps. Let $f_{b,k,j}$ and $\widehat f_{b,k,j}^{(\tau)}$ denote standardized demonstrated and decoded finger commands for batch element $b$, step $k$, and finger joint $j$. Define $\Delta f_{b,k,j}=f_{b,k+1,j}-f_{b,k,j}$, with $\Delta\widehat f_{b,k,j}^{(\tau)}$ defined analogously. The finger-difference loss is
\begin{equation}
    \mathcal{L}_{\Delta f}
    =\operatorname*{mean}_{(b,k,j)\,\mathrm{valid}}
      \rho_1\!\left(
        \frac{
          \Delta\widehat f_{b,k,j}^{(\tau)}
          -\Delta f_{b,k,j}
        }{\bar s_j}
      \right),
    \label{eq:finger-difference-loss}
\end{equation}
where $\rho_1$ is the Huber loss with threshold one. The scale $\bar s_j$ is the stopped-gradient, batch-level RMS of demonstrated adjacent differences for joint $j$, subject to a fixed minimum. Both steps of an adjacent pair must satisfy the Effect-valid mask. This loss supervises finger opening and closing without treating joints with different difference magnitudes as equally scaled.

\paragraph{Joint optimization and closed-loop execution.} Stage~II minimizes
\begin{equation}
    \mathcal{L}_{\mathrm{II}}
    =\mathcal{L}_{\mathrm{FM}}
     +r(u)\,\mathbb{E}_{\tau}\!\left[
        g(\tau)
        \bigl(
          \lambda_{\mathrm{GT}}\mathcal{L}_{\mathrm{GT}}
          +\lambda_{\Delta f}\mathcal{L}_{\Delta f}
        \bigr)
      \right],
    \label{eq:stage2-loss}
\end{equation}
where $u$ is the optimization step, $r(u)\in[0,1]$ warms up both action losses, and $g(\tau)$ selects flow times at which the clean-endpoint estimate is supervised. Flow matching remains active over the full training distribution. The gate also accounts for the probability mass of its selected flow-time interval. The ground-truth-action weight is calibrated with reference to its gradient scale relative to flow matching. Flow-time sampling, gating, coefficients, and optimization schedules are provided in Sec.~\ref{app:reproducibility}.

At deployment, the VLA predicts $\widehat{\mathbf{Z}}_t$ from the current observation, $\mathcal{N}^{-1}$ recovers the Effect, and the jointly trained decoder produces an action chunk. We execute a prefix, observe again, and replan using the same receding-horizon protocol as the direct-action baseline. The deployed system retains the VLA, fixed normalization statistics, the required state projection, and the action decoder.

\subsection{Cross-Embodiment Factorization}
\label{sec:cross-embodiment}

Single- and dual-arm demonstrations use different native state and action dimensions. AFCE uses separate input and output adapters while keeping a common Effect shape. For embodiment or control configuration $e$,
\begin{equation}
    \mathbf{h}^{A,e}=f_A^{e}(\mathbf{A}^{e}),
    \qquad
    \mathbf{h}^{S,e}=f_S^{e}(\Delta\mathbf{S}^{e}),
    \qquad
    \widehat{\mathbf{A}}^{e}
    =D_{\psi}^{e}\!\left(
        \widehat{\mathbf{E}},\mathbf{s}_t^{e}
      \right).
    \label{eq:embodiment-adapters}
\end{equation}
Action and response adapters map native coordinates to a common feature width. State projections and action-decoding interfaces are configuration-specific, whereas the visual encoder, cross-modal fusion, temporal Effect modeling, and Effect-prediction backbone are shared. The matching decoder $D_{\psi}^{e}$ is used for both Stage~II supervision and deployment.

For configuration-specific demonstration distributions $\mathcal{D}_j^{e}$, mixed training at either stage $j\in\{\mathrm{I},\mathrm{II}\}$ can be written as
\begin{equation}
    \mathcal{L}_{j}^{\mathrm{mix}}
    =\sum_{e\in\mathcal{E}}\omega_e^{(j)}
      \mathbb{E}_{\xi\sim\mathcal{D}_j^{e}}
      \left[\mathcal{L}_{j}^{e}(\xi)\right],
    \qquad
    \omega_e^{(j)}\geq0,
    \quad
    \sum_{e\in\mathcal{E}}\omega_e^{(j)}=1.
    \label{eq:mixed-embodiment-loss}
\end{equation}
The weights describe the mixture induced by the training sampler. In Stage~II, sampling is proportional to eligible frame counts rather than assigning equal weight to each task. Reconstruction losses are averaged over valid coordinates within each configuration, preventing larger native action spaces from receiving greater weight solely due to their dimensionality. This factorization enables heterogeneous demonstrations to train a shared Effect representation without a common native action space. It does not require different robots to produce identical Effect codes or follow identical trajectories for similar outcomes; embodiment-invariant alignment would require separate evidence.

\raggedbottom
\setlength{\intextsep}{6pt plus 2pt minus 2pt}
\section{Experiments}
\label{sec:experiments}

\subsection{Experimental Setup}
\label{sec:experimental-setup}

\paragraph{Benchmark and compared methods.} We evaluate on DexJoCo~\citep{wang2026dexjoco}, which comprises six single-arm and five bimanual dexterous manipulation tasks. All reported evaluations use \texttt{rand-obj}, with randomized object placement and table height. We compare our method with UVT~\citep{feng2026unifiedvisuomotortargetssupervising} and the officially released DexJoCo $\pi_{0.5}$ multi-task checkpoint, denoted as $\pi_{0.5}$ in Table~\ref{tab:dexjoco-main}. Inputs comprise scene and wrist images, the language instruction, and proprioceptive state.

\paragraph{Evaluation protocol.} Each method is evaluated for 50 episodes per task under each of three environment seeds (0, 1, and 2), totaling 1,650 episodes per method. We report mean success rate and sample standard deviation across these \emph{evaluation} seeds, using one trained checkpoint per method. These statistics describe variation across evaluation seeds, not independent training runs. For each group, we first average task success rates within each seed and then compute the mean and sample standard deviation across seeds. The overall score averages all 11 tasks equally.

\begin{table}[h]
    \centering
    \vspace{-10pt}
    \caption{Success rates (\%) on DexJoCo under \texttt{rand-obj}.}
    \label{tab:dexjoco-main}
    \fontsize{7.5}{9}\selectfont
    \renewcommand{\tablesr}[2]{\shortstack[c]{$#1$\\[1pt]{\fontsize{6.5}{7.5}\selectfont$\pm #2$}}}
    \setlength{\tabcolsep}{1pt}
    \renewcommand{\arraystretch}{1.25}
    \arrayrulecolor{tableink}
    \begin{tabularx}{\linewidth}{l*{14}{>{\centering\arraybackslash}X}}
        \toprule[0.9pt]
        & \multicolumn{6}{c}{\cellcolor{tablemint}\textbf{Single-arm}}
        & \multicolumn{5}{c}{\cellcolor{tablelavender}\textbf{Bimanual}}
        & \multicolumn{3}{c}{\cellcolor{tableblue}\textbf{Mean}} \\
        \cmidrule(lr){2-7}\cmidrule(lr){8-12}\cmidrule(lr){13-15}
        \textbf{Method} & \textbf{PT} & \textbf{CM} & \textbf{FG} & \textbf{HN}
        & \textbf{PB} & \textbf{WP} & \textbf{UI} & \textbf{HA} & \textbf{AS}
        & \textbf{MW} & \textbf{PH} & \textbf{SA} & \textbf{BA} & \textbf{All} \\
        \midrule[0.45pt]
        UVT & \tablesr{1.33}{1.15} & \tablesr{\mathbf{80.00}}{2.00} & \tablesr{\mathbf{64.67}}{5.03} & \tablesr{22.00}{6.93} & \tablesr{60.00}{6.00}
        & \tablesr{\mathbf{92.67}}{2.31} & \tablesr{0.00}{0.00} & \tablesr{0.67}{1.15} & \tablesr{0.00}{0.00} & \tablesr{41.33}{11.02}
        & \tablesr{30.67}{4.16} & \tablesr{53.44}{2.22} & \tablesr{14.53}{3.06} & \tablesr{35.76}{0.56} \\
        \addlinespace[3pt]
        $\pi_{0.5}$ & \tablesr{\mathbf{43.33}}{8.08} & \tablesr{60.00}{2.00} & \tablesr{38.00}{5.29} & \tablesr{84.67}{5.77} & \tablesr{67.33}{1.15}
        & \tablesr{88.67}{1.15} & \tablesr{2.00}{0.00} & \tablesr{5.33}{4.16} & \tablesr{1.33}{2.31} & \tablesr{32.00}{8.72}
        & \tablesr{8.00}{3.46} & \tablesr{\mathbf{63.67}}{1.86}
        & \tablesr{9.73}{2.41} & \tablesr{39.15}{2.11} \\
        \addlinespace[3pt]
        \rowcolor{tablecream}
        \textbf{Ours} & \tablesr{5.33}{2.31} & \tablesr{54.67}{7.57} & \tablesr{62.00}{6.93} & \tablesr{\mathbf{86.67}}{5.77} & \tablesr{\mathbf{80.67}}{9.02}
        & \tablesr{84.67}{9.02} & \tablesr{\mathbf{7.33}}{1.15}
        & \tablesr{\mathbf{29.33}}{1.15} & \tablesr{\mathbf{9.33}}{3.06}
        & \tablesr{\mathbf{81.33}}{6.43}
        & \tablesr{\mathbf{59.33}}{4.16} & \tablesr{62.33}{1.45} & \tablesr{\mathbf{37.33}}{2.66} & \tablesr{\mathbf{50.97}}{1.65} \\
        \bottomrule[0.9pt]
    \end{tabularx}
    \par\vspace{3pt}
    \begin{minipage}{\linewidth}
        \footnotesize
        Mean $\pm$ sample standard deviation across three environment seeds;
        50 episodes per task per seed. Bold denotes the highest mean among
        reported methods.\\
        PT: Pinch Tongs; CM: Click Mouse; FG: Fold Glasses; HN: Hammer Nail;
        PB: Pick Bucket; WP: Water Plant; UI: Unlock iPad; HA: Hanoi;
        AS: Assembly; MW: Microwave; PH: Photograph.
        SA/BA/All: single-arm, bimanual, and overall task means.
    \end{minipage}
    \vspace{-10pt}
\end{table}

\subsection{Main Results}
\label{sec:main-results}

\paragraph{Overall performance.} Ours achieves $50.97\pm1.65\%$ (841/1,650 episodes), exceeding $\pi_{0.5}$ by 11.82 percentage points ($39.15\pm2.11\%$) and UVT by 15.21 points ($35.76\pm0.56\%$). It obtains the highest mean on seven tasks.

\paragraph{Bimanual manipulation.} Ours leads on all five bimanual tasks, reaching a mean of $37.33\%$, compared with $9.73\%$ for $\pi_{0.5}$ and $14.53\%$ for UVT. The largest gains over $\pi_{0.5}$ occur on Microwave ($81.33\%$ versus $32.00\%$) and Photograph ($59.33\%$ versus $8.00\%$). Success on Hanoi also rises from $5.33\%$ to $29.33\%$. Ours improves on Assembly and Unlock iPad, although their success rates remain low at $9.33\%$ and $7.33\%$, respectively.

\begin{wrapfigure}[14]{r}{0.55\textwidth}
    \centering
    \vspace{-8pt}
    \includegraphics[width=\linewidth]{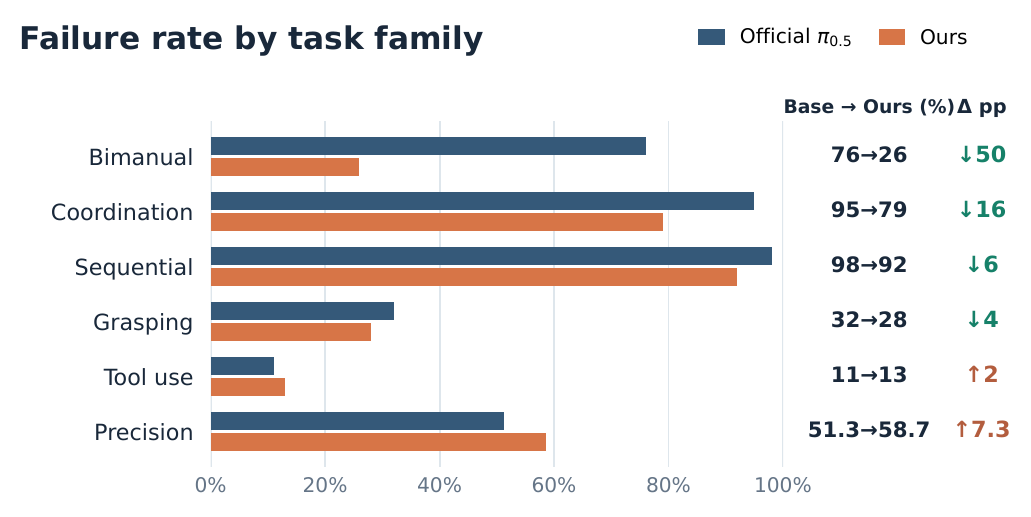}
    \caption{Failure rates by task family for evaluation seed 0.
    Arrows show changes from $\pi_{0.5}$ to Ours in percentage points.}
    \label{fig:dexjoco-family-failure}
    \vspace{-8pt}
\end{wrapfigure}

\paragraph{Single-arm manipulation.} Ours achieves a single-arm mean of $62.33\%$, slightly below $\pi_{0.5}$ ($63.67\%$) but above UVT ($53.44\%$). It leads on Hammer Nail and Pick Bucket and improves over $\pi_{0.5}$ on Fold Glasses ($62.00\%$ versus $38.00\%$), although UVT performs best on the latter task. Pinch Tongs is the largest regression: Ours reaches $5.33\%$, versus $43.33\%$ for $\pi_{0.5}$. The seed-0 precision-task failure rate in Figure~\ref{fig:dexjoco-family-failure} likewise rises from $51.3\%$ to $58.7\%$. Thus, the overall gain does not extend uniformly to fine-grained single-arm control.

\paragraph{Source of the aggregate gain.} With equal numbers of evaluation episodes per task, the 27.60-point bimanual gain over $\pi_{0.5}$ contributes approximately 12.55 points to the overall mean; the 1.34-point single-arm deficit offsets approximately 0.73 points. The resulting net gain is 11.82 points.

Figure~\ref{fig:dexjoco-family-failure} groups the tasks into Tool use (Hammer Nail, Water Plant), Grasping (Pick Bucket), Precision (Pinch Tongs, Click Mouse, Fold Glasses), Bimanual (Microwave, Photograph), Coordination (Hanoi, Assembly), and Sequential (Unlock iPad). The groups are disjoint; in particular, the figure's Bimanual group contains two tasks, whereas the BA column of Table~\ref{tab:dexjoco-main} averages all five bimanual tasks.

On seed 0, Ours reduces the failure rate from $76\%$ to $26\%$ in Bimanual and from $95\%$ to $79\%$ in Coordination. Sequential improves from $98\%$ to $92\%$, but Unlock iPad remains difficult. Failure rates instead rise in Precision ($51.3\%$ to $58.7\%$) and Tool use ($11\%$ to $13\%$). This analysis localizes gains and regressions of the complete policies; the ablations examine the contribution of Effect-construction inputs.

\subsection{Effect Dimensionality Sensitivity}
\label{sec:effect-dim-sensitivity}

We train $30\times d$ Effects with $d\in\{256,128,64\}$ for 30,000 steps under identical settings. Table~\ref{tab:effect-dim-sensitivity} reports decoding errors and GT-Effect replay on the same selected set of 110 training demonstrations, with 10 demonstrations per task. This diagnostic evaluates the Effect and decoder without policy-side Effect prediction and is reported separately from closed-loop policy success.

\begin{table}[H]
    \centering
    \vspace{-10pt}
    \caption{Effect dimensionality sensitivity.}
    \label{tab:effect-dim-sensitivity}
    \fontsize{8}{9.5}\selectfont
    \setlength{\tabcolsep}{2.5pt}
    \renewcommand{\arraystretch}{1.3}
    \arrayrulecolor{tableink}
    \begin{tabularx}{\linewidth}
        {l*{6}{>{\centering\arraybackslash}X}}
        \toprule[0.9pt]
        & \multicolumn{3}{c}{\cellcolor{tablemint}\textbf{Error (mm) $\downarrow$}}
        & \multicolumn{3}{c}{\cellcolor{tablelavender}\textbf{GT-Effect replay $\uparrow$}} \\
        \cmidrule(lr){2-4}\cmidrule(lr){5-7}
        \textbf{$d$} & \textbf{Pos.} & \textbf{Bi-pos.} & \textbf{FK}
        & \textbf{SA} & \textbf{BA} & \textbf{All} \\
        \midrule[0.45pt]
        \rowcolor{tablecream}
        \textbf{256} & \textbf{7.794} & \textbf{9.507} & \textbf{10.519}
        & 20/60 & \textbf{5/50} & \textbf{25/110} \\
        128 & 8.965 & 11.455 & 11.716
        & \textbf{21/60} & 2/50 & 23/110 \\
        64 & 11.585 & 15.272 & 14.980
        & 18/60 & 0/50 & 18/110 \\
        \bottomrule[0.9pt]
    \end{tabularx}
    \par\vspace{3pt}
    \begin{minipage}{\linewidth}
        \footnotesize
        Pos.: overall position error; Bi-pos.: bimanual position error; FK: fingertip
        forward-kinematics position error. SA/BA/All: successful single-arm, bimanual, and
        total replay episodes. Bold indicates the best value in each column.
    \end{minipage}
\end{table}

Reducing $d$ from 256 to 64 increases all three decoding errors and lowers total GT-Effect replay success from 25/110 to 18/110. The trend differs slightly by task group: single-arm replay rises from 20/60 to 21/60 at $d=128$, but bimanual replay falls from 5/50 to 2/50 and reaches 0/50 at $d=64$. We therefore retain $d=256$, which gives the lowest measured reconstruction errors and highest total replay success among these variants. The modest absolute replay rate also cautions against treating representation reconstruction alone as evidence of reliable execution.

\subsection{Effect-Construction Ablation}
\label{sec:ablations}

\begin{table}[H]
    \centering
    \vspace{-10pt}
    \caption{Effect-construction input ablation on DexJoCo under
    \texttt{rand-obj}. Success rates are reported in \%.}
    \label{tab:effect-input-ablation}
    \fontsize{7.5}{9}\selectfont
    \newcommand{\ablval}[2]{%
        \shortstack[c]{$#1$\\[1pt]%
        {\fontsize{6.5}{7.5}\selectfont$\pm #2$}}%
    }
    \setlength{\tabcolsep}{1.5pt}
    \renewcommand{\arraystretch}{1.2}
    \arrayrulecolor{tableink}

    \begin{tabularx}{\linewidth}{@{}lccc*{7}{>{\centering\arraybackslash}X}@{}}
        \toprule[0.9pt]
        \textbf{Variant}
        & \multicolumn{3}{c}{\textbf{Effect inputs}}
        & \multicolumn{7}{c}{\cellcolor{tablemint}\textbf{Single-arm}} \\
        \cmidrule(lr){2-4}\cmidrule(lr){5-11}
        & \textbf{$S$} & \textbf{$V$} & \textbf{$A$}
        & \textbf{PT} & \textbf{CM} & \textbf{FG}
        & \textbf{HN} & \textbf{PB} & \textbf{WP}
        & \textbf{SA} \\
        \midrule[0.45pt]

        $A$ only
        & -- & -- & $\checkmark$
        & \ablval{1.33}{1.15}
        & \ablval{65.33}{3.06}
        & \ablval{54.67}{10.07}
        & \ablval{68.00}{2.00}
        & \ablval{69.33}{6.11}
        & \ablval{84.67}{2.31}
        & \ablval{57.22}{0.38} \\

        w/o $S$
        & -- & $\checkmark$ & $\checkmark$
        & \ablval{5.33}{5.03}
        & \ablval{48.00}{5.29}
        & \ablval{51.33}{4.62}
        & \ablval{76.00}{2.00}
        & \ablval{84.67}{7.02}
        & \ablval{26.67}{5.03}
        & \ablval{48.67}{1.53} \\

        w/o $V$
        & $\checkmark$ & -- & $\checkmark$
        & \ablval{14.00}{5.29}
        & \ablval{57.33}{3.06}
        & \ablval{54.67}{6.43}
        & \ablval{78.67}{8.08}
        & \ablval{76.67}{9.02}
        & \ablval{52.67}{12.06}
        & \ablval{55.67}{3.06} \\

        \rowcolor{tablecream}
        \textbf{AFCE}
        & $\checkmark$ & $\checkmark$ & $\checkmark$
        & \ablval{5.33}{2.31}
        & \ablval{54.67}{7.57}
        & \ablval{62.00}{6.93}
        & \ablval{86.67}{5.77}
        & \ablval{80.67}{9.02}
        & \ablval{84.67}{9.02}
        & \ablval{\mathbf{62.33}}{1.45} \\
        \bottomrule[0.9pt]
    \end{tabularx}

    \par\vspace{6pt}

    \begin{tabularx}{\linewidth}{@{}lccc*{6}{>{\centering\arraybackslash}X}@{}}
        \toprule[0.9pt]
        \textbf{Variant}
        & \multicolumn{3}{c}{\textbf{Effect inputs}}
        & \multicolumn{6}{c}{\cellcolor{tablelavender}\textbf{Bimanual}} \\
        \cmidrule(lr){2-4}\cmidrule(lr){5-10}
        & \textbf{$S$} & \textbf{$V$} & \textbf{$A$}
        & \textbf{UI} & \textbf{HA} & \textbf{AS}
        & \textbf{MW} & \textbf{PH} & \textbf{BA} \\
        \midrule[0.45pt]

        $A$ only
        & -- & -- & $\checkmark$
        & \ablval{2.67}{1.15}
        & \ablval{28.67}{8.33}
        & \ablval{4.00}{5.29}
        & \ablval{76.00}{2.00}
        & \ablval{43.33}{7.57}
        & \ablval{30.93}{2.81} \\

        w/o $S$
        & -- & $\checkmark$ & $\checkmark$
        & \ablval{3.33}{1.15}
        & \ablval{11.33}{4.16}
        & \ablval{1.33}{1.15}
        & \ablval{57.33}{9.87}
        & \ablval{54.00}{6.93}
        & \ablval{25.47}{2.66} \\

        w/o $V$
        & $\checkmark$ & -- & $\checkmark$
        & \ablval{1.33}{1.15}
        & \ablval{34.00}{3.46}
        & \ablval{2.00}{0.00}
        & \ablval{61.33}{5.03}
        & \ablval{9.33}{4.62}
        & \ablval{21.60}{1.20} \\

        \rowcolor{tablecream}
        \textbf{AFCE}
        & $\checkmark$ & $\checkmark$ & $\checkmark$
        & \ablval{7.33}{1.15}
        & \ablval{29.33}{1.15}
        & \ablval{9.33}{3.06}
        & \ablval{81.33}{6.43}
        & \ablval{59.33}{4.16}
        & \ablval{\mathbf{37.33}}{2.66} \\
        \bottomrule[0.9pt]
    \end{tabularx}

    \par\vspace{3pt}
    \begin{minipage}{\linewidth}
        \footnotesize
        A checkmark indicates evidence used to construct the Effect target;
        -- indicates its removal. $S$: realized robot response;
        $V$: future world transition; $A$: executed actions.
        SA and BA denote the single-arm and bimanual task means.
        Values are mean $\pm$ sample standard deviation across three
        environment seeds, with 50 episodes per task per seed.
        Task abbreviations follow Table~\ref{tab:dexjoco-main}.
    \end{minipage}
\end{table}

Table~\ref{tab:effect-input-ablation} varies the evidence supplied to the Effect tokenizer while retaining executed actions as an input. With only $A$ as input, the policy achieves $57.22\%$ success on single-arm tasks and $30.93\%$ on bimanual tasks. Adding $V$ alone yields $48.67\%$ and $25.47\%$, while adding $S$ alone yields $55.67\%$ and $21.60\%$. The complete $A+S+V$ configuration achieves the highest group means, $62.33\%$ and $37.33\%$, exceeding the $A$-only variant by $5.11$ and $6.40$ percentage points.

At the task level, full AFCE improves Hammer Nail from $68.00\%$ to $86.67\%$ and Photograph from $43.33\%$ to $59.33\%$ relative to $A$ only, but reduces Click Mouse from $65.33\%$ to $54.67\%$. The nonmonotonic results support the complete configuration at the group level without establishing an independent benefit from each added input; closed-loop success alone cannot distinguish errors in Effect prediction, fusion, and action decoding.

\section{Limitations}
\label{sec:limitations}

AFCE requires temporally aligned actions, robot states, and future images to construct training targets. Visual features can miss subtle contacts or be affected by occlusion and viewpoint changes. A fixed-size Effect must also preserve control detail while remaining predictable from current observations; the Pinch Tongs result illustrates an unresolved precision-control limitation. Prediction errors can further degrade action readout, and offline target construction increases training cost. Finally, shared parameters across single- and bimanual interfaces do not establish embodiment-invariant representations.

\section{Conclusion}
\label{sec:conclusion}

We introduced AFCE, which constructs a temporal Effect from demonstrated actions, realized robot responses, and sparse visual transitions, then predicts and decodes it for closed-loop control. On DexJoCo under \texttt{rand-obj}, AFCE achieves 50.97\% overall success, compared with 39.15\% for $\pi_{0.5}$ and 35.76\% for the evaluated UVT baseline. The gains are strongest on bimanual tasks, while precision-sensitive single-arm control remains challenging. AFCE uses future physical evidence for supervision while requiring only current observations at inference time.








\FloatBarrier
\bibliographystyle{iclr2027_conference}
\bibliography{iclr2027_conference}


\newpage
\appendix
\numberwithin{figure}{section}
\numberwithin{table}{section}
\numberwithin{equation}{section}

\section{Method and Objective Details}
\label{app:method-details}

AFCE separates effect construction from effect prediction. Given a demonstration window beginning at time $t$, the first stage encodes the executed actions $A_{t:t+H-1}$, realized robot responses $\Delta S_{t:t+H-1}$, and future visual transitions $V_t$ into an Effect target $E_t^*$. The second stage predicts this target from the current image, instruction, and robot state, and decodes the predicted Effect into an executable action chunk. Demonstration futures are used to construct training targets but are not inputs to the deployed policy.

\subsection{Effect Evidence and Architecture}

The action branch represents the demonstrated control sequence. The response branch describes realized changes in palm pose, fingertip positions, and finger joints relative to the beginning of the window. The visual branch extracts sparse transition descriptors from future observations using a frozen visual encoder. These descriptors may include robot motion; the visual stream is not restricted to object-only changes.

Each branch produces temporally aligned features, which are projected to a shared width and augmented with modality embeddings. At each future step, a learned query attends to the action, response, and visual features through QueryFusion. A temporal Transformer then exchanges information across the window to produce
\begin{equation}
    E_t^*
    =T_\phi\!\left(
        A_{t:t+H-1},
        \Delta S_{t:t+H-1},
        V_t
      \right)
    \in\mathbb{R}^{H\times d_E}.
    \label{eq:app-effect-tokenizer}
\end{equation}
A state-conditioned decoder $D_\psi^{e}$ maps the Effect to an action sequence in the native coordinates of embodiment $e$. Separate output interfaces accommodate different action dimensions while the Effect representation has a common shape.

\subsection{Learning an Executable Effect}

The tokenizer is trained with action reconstruction and auxiliary response and visual reconstruction. Let $\mathcal H_p$, $\mathcal H_r$, and $\mathcal H_f$ denote grouped Huber errors for standardized position, rotation, and finger actions, averaged over valid steps and coordinates. The action loss is
\begin{equation}
    \mathcal L_A
    =\frac{w_p\mathcal H_p+\mathcal H_r+\mathcal H_f}{3},
    \label{eq:app-action-reconstruction}
\end{equation}
where $w_p$ controls the emphasis on position reconstruction. The denominator remains three when $w_p\neq1$. We additionally measure the discrepancy between decoded and demonstrated orientations by the SO(3) chordal loss
\begin{equation}
    \mathcal L_{\mathrm{SO(3)}}
    =\mathbb E\!\left[
        \frac{\|R(\hat A)-R(A)\|_F^2}{6}
      \right].
    \label{eq:app-rotation-reconstruction}
\end{equation}

The response and visual heads reconstruct their respective normalized targets on valid entries, yielding losses $\mathcal L_S$ and $\mathcal L_V$. The complete first-stage objective is
\begin{equation}
    \mathcal L_{\mathrm I}
    =\mathcal L_A
     +\lambda_S(u)\mathcal L_S
     +\lambda_V(u)\mathcal L_V
     +\lambda_R\mathcal L_{\mathrm{SO(3)}},
    \label{eq:app-effect-objective}
\end{equation}
where $u$ is the optimizer step. The auxiliary weights are gradually increased during tokenizer training. Their schedules and the remaining numerical settings are given in Sec.~\ref{app:reproducibility}.

After training, we encode the demonstration windows into fixed Effect targets. Channel-wise statistics fitted on all tokenizer training demonstrations define the normalized policy target
\begin{equation}
    Z_t^*=\frac{E_t^*-\mu_E}{\sigma_E^{\mathrm{safe}}},
    \label{eq:app-effect-standardization}
\end{equation}
where $\sigma_E^{\mathrm{safe}}$ includes a minimum standard-deviation threshold. The tokenizer and normalization statistics remain fixed in the second stage.

\subsection{Effect Prediction and Action Readout}

The AFCE policy is initialized from $\pi_{0.5}$, with its action projections adapted to the Effect dimension. Conditioned on the current observation and robot state, it learns a velocity field in normalized Effect space. For $\epsilon\sim\mathcal N(0,I)$ and flow time $\tau$, we construct
\begin{equation}
    Z_{t,\tau}=(1-\tau)Z_t^*+\tau\epsilon,
    \qquad
    v^*=\epsilon-Z_t^*.
    \label{eq:app-effect-interpolant}
\end{equation}
The flow-matching objective averages squared velocity error over valid Effect coordinates:
\begin{equation}
    \mathcal L_{\mathrm{FM}}
    =\mathbb E_{\tau,\epsilon}
      \left[
        \operatorname{mean}_{k,j}
        \bigl(v_{\theta,k,j}-v^*_{k,j}\bigr)^2
      \right].
    \label{eq:app-effect-flow-loss}
\end{equation}

To connect Effect prediction to executable control during training, we form the single-step clean estimate
\begin{equation}
    \hat Z_{t,0}^{(\tau)}
    =Z_{t,\tau}-\tau v_\theta(Z_{t,\tau},\tau;X_t),
    \qquad
    \hat A_t^{(\tau)}
    =D_\psi^e\!\left(
        \mathcal N_E^{-1}(\hat Z_{t,0}^{(\tau)}),
        S_t
      \right),
    \label{eq:app-effect-readout}
\end{equation}
where $X_t$ is the current image--language--state context and $\mathcal N_E^{-1}$ reverses Effect standardization. The ground-truth-action auxiliary loss $\mathcal L_{\mathrm{GT}}$ compares $\hat A_t^{(\tau)}$ with the standardized demonstrated action. It averages position, rotation, and finger Huber errors with equal group weights; the position multiplier and SO(3) term from Stage I are not applied here.

A second auxiliary loss preserves finger motion within the decoded chunk. For consecutive valid steps, define $\Delta f_{t+k}=f_{t+k+1}-f_{t+k}$ and the corresponding decoded difference $\Delta\hat f_{t+k}$. Each joint's difference error is scaled by the stopped-gradient RMS of its demonstrated differences, subject to a minimum scale, before applying a Huber loss. Averaging over valid adjacent pairs and finger joints gives $\mathcal L_{\Delta f}$.

Both action losses update the policy and the same embodiment decoder used at inference. They share a flow-time gate $g(\tau)$ and a training ramp $r(u)$:
\begin{equation}
    \mathcal L_{\mathrm{II}}
    =\mathcal L_{\mathrm{FM}}
     +r(u)g(\tau)
      \left(
        \lambda_{\mathrm{GT}}\mathcal L_{\mathrm{GT}}
        +\lambda_{\Delta f}\mathcal L_{\Delta f}
      \right).
    \label{eq:app-afce-policy-objective}
\end{equation}
The gate restricts action supervision to a designated flow-time interval and accounts for its probability mass under the sampling distribution. The exact interval, weights, ramp, and optimizer settings are reported in Sec.~\ref{app:reproducibility}. At inference, the policy generates a normalized Effect, which is denormalized and passed through this jointly trained decoder to obtain a native action chunk. The future-evidence tokenizer is not run online.

\section{Datasets and Evaluation Protocols}
\label{app:evaluation-details}

We evaluate AFCE on all 11 DexJoCo tasks \citep{wang2026dexjoco}, comprising six single-arm and five bimanual Panda--Allegro manipulation tasks. These tasks involve tool use, articulated finger motions, ordered interactions, and bimanual coordination. Table~\ref{tab:app-dexjoco-tasks} summarizes their functional goals. Success is determined by each task's simulator-defined completion condition.

\begin{table}[htbp]
    \centering
    \caption{Functional goals of the 11 DexJoCo tasks.}
    \label{tab:app-dexjoco-tasks}
    \footnotesize
    \setlength{\tabcolsep}{4pt}
    \renewcommand{\arraystretch}{1.12}
    \begin{tabularx}{\linewidth}{@{}>{\raggedright\arraybackslash}p{0.26\linewidth}>{\raggedright\arraybackslash}X@{}}
        \toprule
        \textbf{Task} & \textbf{Objective} \\
        \midrule
        \multicolumn{2}{@{}l}{\textit{Single arm}} \\
        Click Mouse & Move the mouse onto its pad and click. \\
        Fold Glasses & Fold the glasses and place them in the case. \\
        Hammer Nail & Drive a nail into a board with a hammer. \\
        Pick Bucket & Place food in the bucket and lift it. \\
        Pinch Tongs & Grasp the tongs and open and close them three times. \\
        Water Plant & Use a watering can to water the plant. \\
        \midrule
        \multicolumn{2}{@{}l}{\textit{Bimanual}} \\
        Assembly & Hold a tray and insert a peg into its hole. \\
        Hanoi & Complete the final two moves of a three-level Tower of Hanoi. \\
        Microwave & Place food inside, close the door, and press start. \\
        Photograph & Aim a camera and press its shutter. \\
        Unlock iPad & Hold the device and enter the password 123. \\
        \bottomrule
    \end{tabularx}
\end{table}

We use all 100 recorded episodes per task, totaling 1,100 episodes and 523,763 retained frames at 30~FPS, to train both the Stage-I Effect tokenizer and the Stage-II policy. Effect targets are constructed offline from demonstration actions, robot responses, and future visual transitions. During rollout, the policy receives only the current image observation, language instruction, and robot proprioception; future evidence and privileged simulator object states are unavailable. Each arm's action contains end-effector position (3), rotation vector (3), and finger joints (16), giving 22 action dimensions for single-arm tasks and 44 for bimanual tasks. The corresponding proprioceptive state has 23 or 46 dimensions. The visual-transition pipeline uses the ego view for bimanual tasks, \texttt{ego\_right} for Click Mouse, and the front view for the remaining single-arm tasks.

Closed-loop evaluation uses the DexJoCo \texttt{rand-obj} setting, which randomizes object placement and table height. The results in our main table aggregate 150 rollouts per task, giving 1,650 rollouts per method across all 11 tasks. At each policy query, ten flow denoising steps produce a 30-step Effect chunk; the jointly trained decoder converts it into actions for execution and subsequent replanning. An episode succeeds when the benchmark's task-specific completion condition is met. If $n_i$ denotes the number of successful rollouts on task $i$, we report $\mathrm{SR}_i=100n_i/150$ and $\mathrm{SR}_{\mathrm{all}}=100\sum_{i=1}^{11}n_i/1650$. The single-arm and bimanual aggregates comprise 900 and 750 rollouts, respectively. AFCE achieves 841/1,650 overall successes (50.97\%), including 561/900 (62.33\%) on single-arm tasks and 280/750 (37.33\%) on bimanual tasks. Reconstruction and GT-Effect replay on the demonstrations are reported as training-data diagnostics, separately from closed-loop policy success.

\section{Additional Experimental Results}
\label{app:additional-results}

We provide complementary diagnostics of the Effect representation, its action readout, and the sparse visual evidence used to construct it. Unless stated otherwise, the analyses use the frozen AFCE Effect tokenizer at update 30,000 and the policy checkpoint at update 59,999. The 110 diagnostic trajectories (ten per task) belong to the demonstrations used to train both the Stage-I tokenizer and the Stage-II policy. These analyses are in-sample diagnostics, not held-out validation or closed-loop policy evaluations.

For the action-readout diagnostic, $D_0$ denotes the frozen Stage-I decoder, whereas $D_1$ is the jointly adapted Stage-II decoder used at deployment.

\begin{figure}[htbp]
  \centering
  \includegraphics[width=\linewidth]{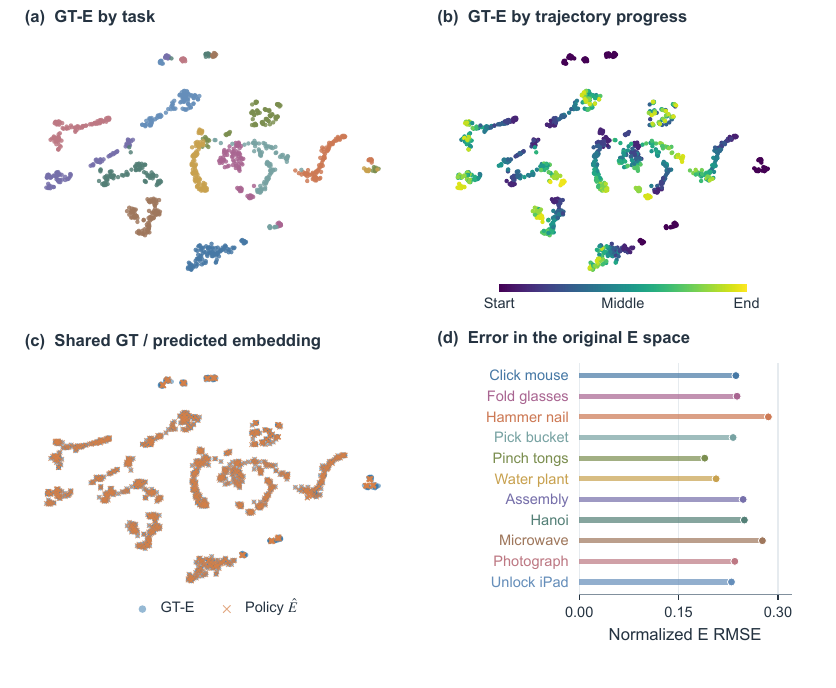}
  \caption{Structure and prediction error of the effect space. (a) GT-E by task; (b) the
  same GT-E windows by relative trajectory progress; (c) GT-E and predicted effects in a
  shared embedding; (d) paired normalized effect RMSE in the original space. PCA uses 50
  components (87.27\% explained variance); the shared t-SNE uses perplexity 30 and seed
  42.}
  \label{fig:app-effect-structure}
\end{figure}

\begin{figure}[htbp]
  \centering
  \includegraphics[width=\linewidth]{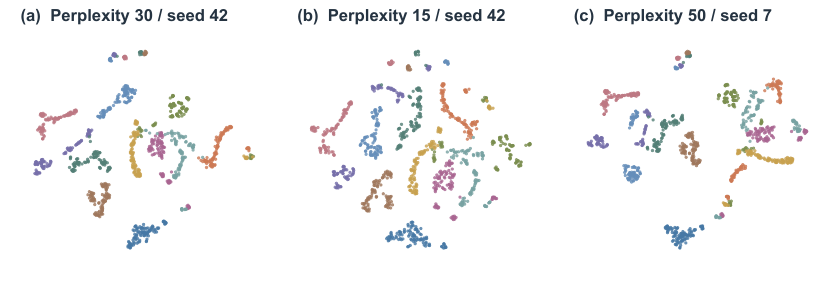}
  \caption{Sensitivity of the two-dimensional projection. GT-E windows after the same
  normalization and PCA are projected with t-SNE at perplexity/seed 30/42, 15/42, and
  50/7. Task colors match Figure~\ref{fig:app-effect-structure}.}
  \label{fig:app-tsne-sensitivity}
\end{figure}

\begin{figure}[htbp]
  \centering
  \includegraphics[width=\linewidth]{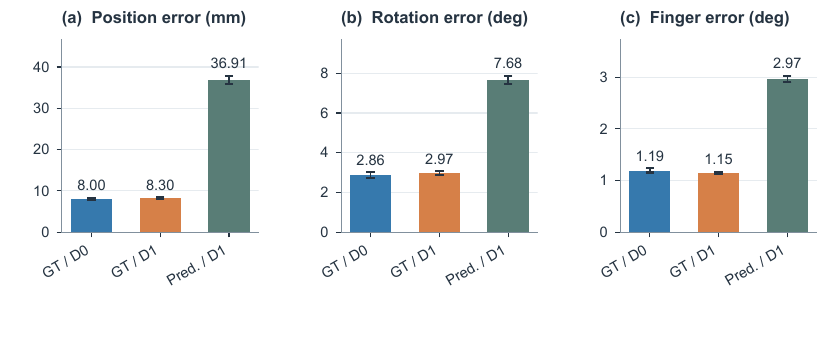}
  \caption{Physical action-readout error on matched demonstration windows. GT-E/$D_0$,
  GT-E/$D_1$, and predicted-E/$D_1$ are compared on mean Euclidean position error per
  hand, rotation geodesic error, and mean absolute finger joint-angle error. Means
  aggregate ten windows per trajectory over 110 trajectories; intervals are 95\%
  within-task episode-bootstrap intervals (3,000 resamples). $D_0$ is used only for this
  diagnostic; the deployed policy uses $D_1$.}
  \label{fig:app-action-readout}
\end{figure}

\paragraph{Effect-space structure and prediction.} Figure~\ref{fig:app-effect-structure} displays a joint PCA--t-SNE projection of 1,100 GT-E windows and 1,100 corresponding policy-predicted windows. Each window contains $30\times256$ effect values. Task and trajectory-progress colors reveal visible organization in the GT-E points; the fourth panel reports paired prediction error in the original normalized effect space rather than in the two-dimensional projection. Overlap of projected GT-E and predicted effects alone does not establish accurate action readout: the matched position error rises from 8.30 mm with GT-E/$D_1$ to 36.91 mm with predicted-E/$D_1$ (Figure~\ref{fig:app-action-readout}). Figure~\ref{fig:app-tsne-sensitivity} shows that the apparent two-dimensional geometry also depends on t-SNE settings.

\paragraph{Projection sensitivity and physical readout error.} Figure~\ref{fig:app-tsne-sensitivity} compares three prespecified t-SNE settings for the same GT-E windows and fixed PCA representation. The changing layout warns against reading inter-cluster distances as quantitative physical or semantic distances. Figure~\ref{fig:app-action-readout} compares action readout on matched demonstration windows: GT-E/$D_0$ and GT-E/$D_1$ produce position errors of 8.00 and 8.30 mm, respectively, while predicted-E/$D_1$ yields 36.91 mm. Rotation and finger errors likewise rise for predicted-E/$D_1$ (7.68 degrees and 2.97 degrees). These are errors against demonstration actions, not closed-loop task success rates.

\begin{figure*}[p]
  \centering
  \includegraphics[width=\textwidth,height=0.215\textheight,keepaspectratio]{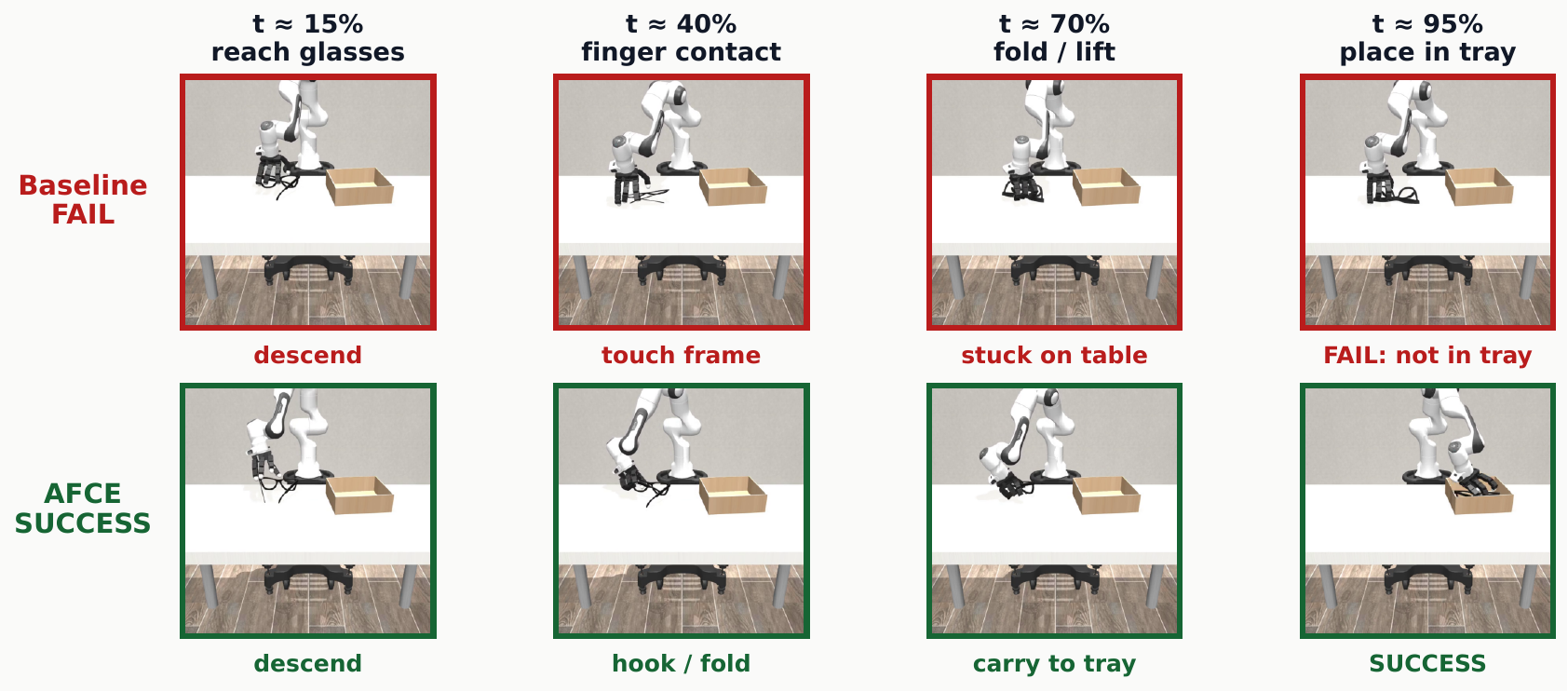}
  \par\vspace{0.006\textheight}
  \includegraphics[width=\textwidth,height=0.215\textheight,keepaspectratio]{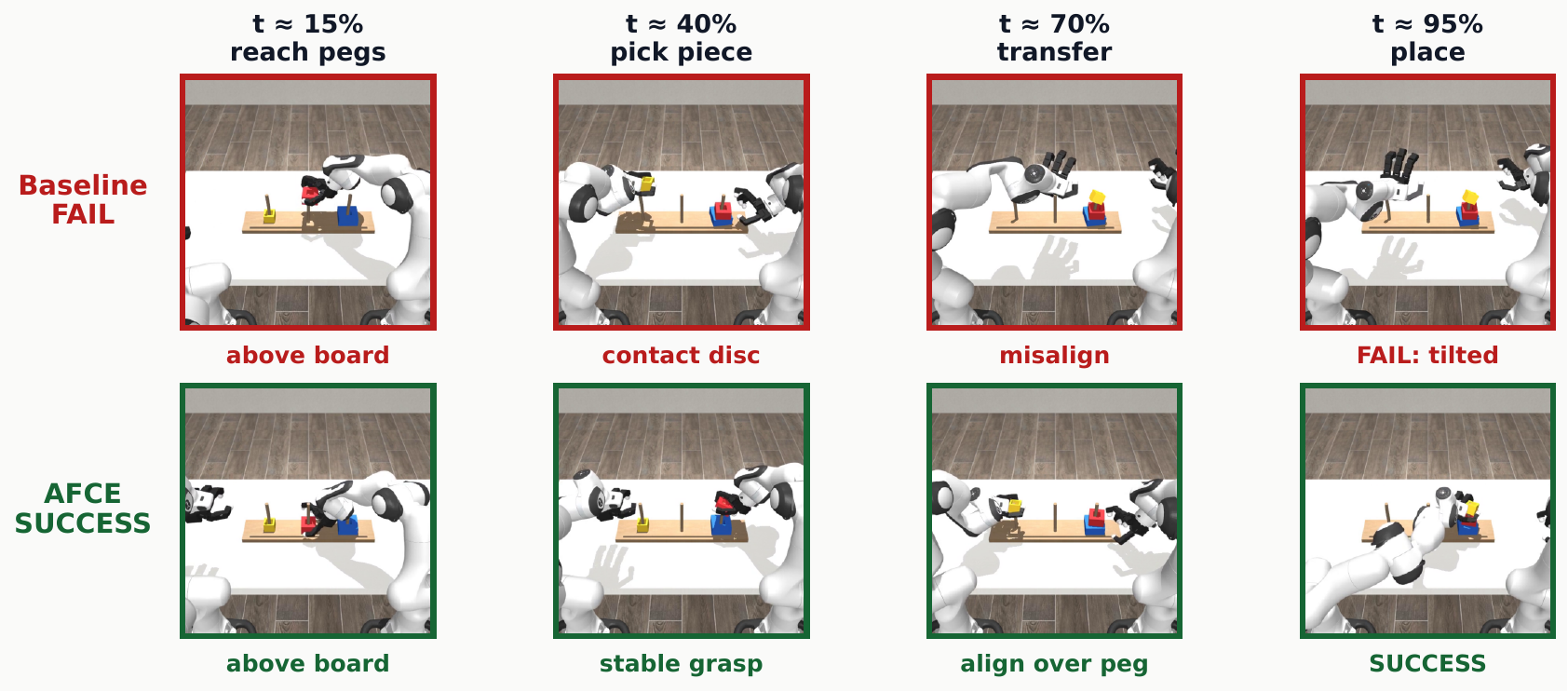}
  \par\vspace{0.006\textheight}
  \includegraphics[width=\textwidth,height=0.215\textheight,keepaspectratio]{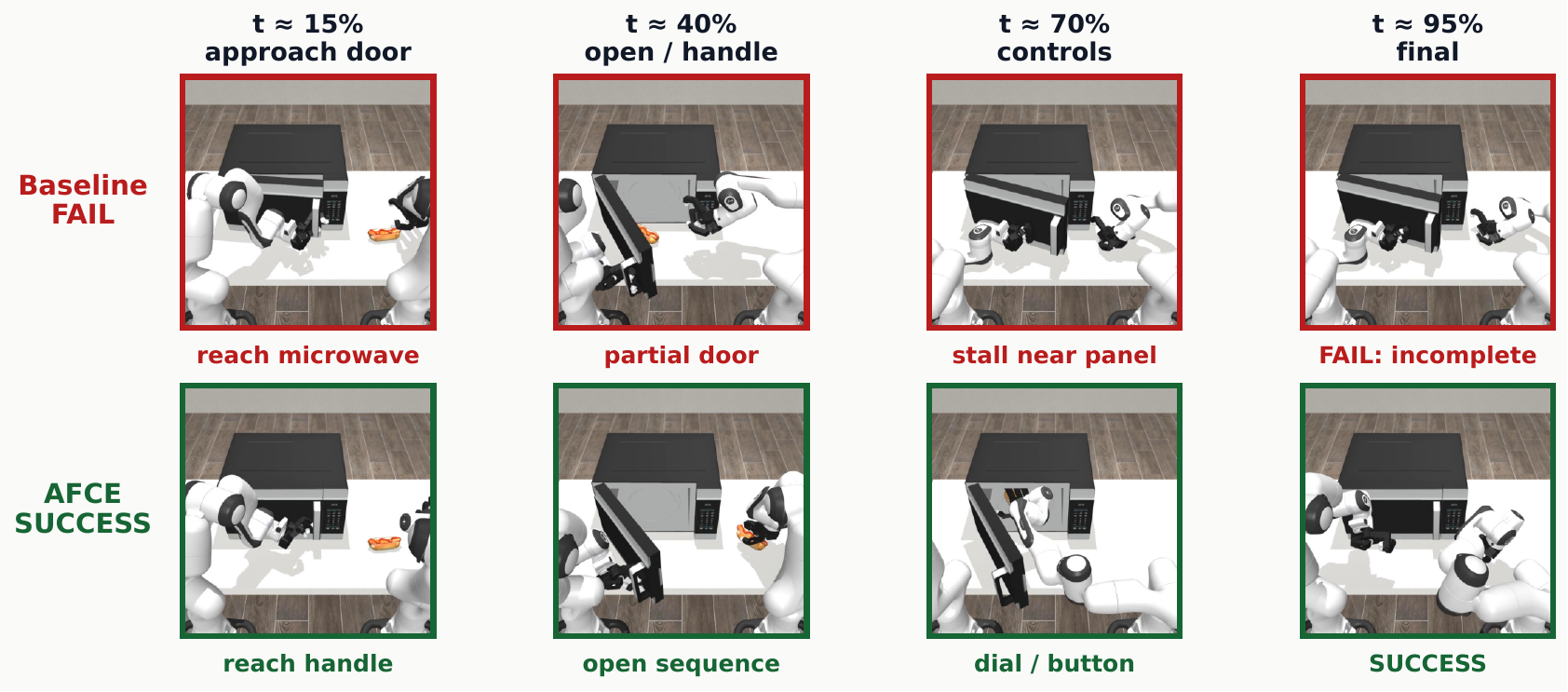}
  \par\vspace{0.006\textheight}
  \includegraphics[width=\textwidth,height=0.215\textheight,keepaspectratio]{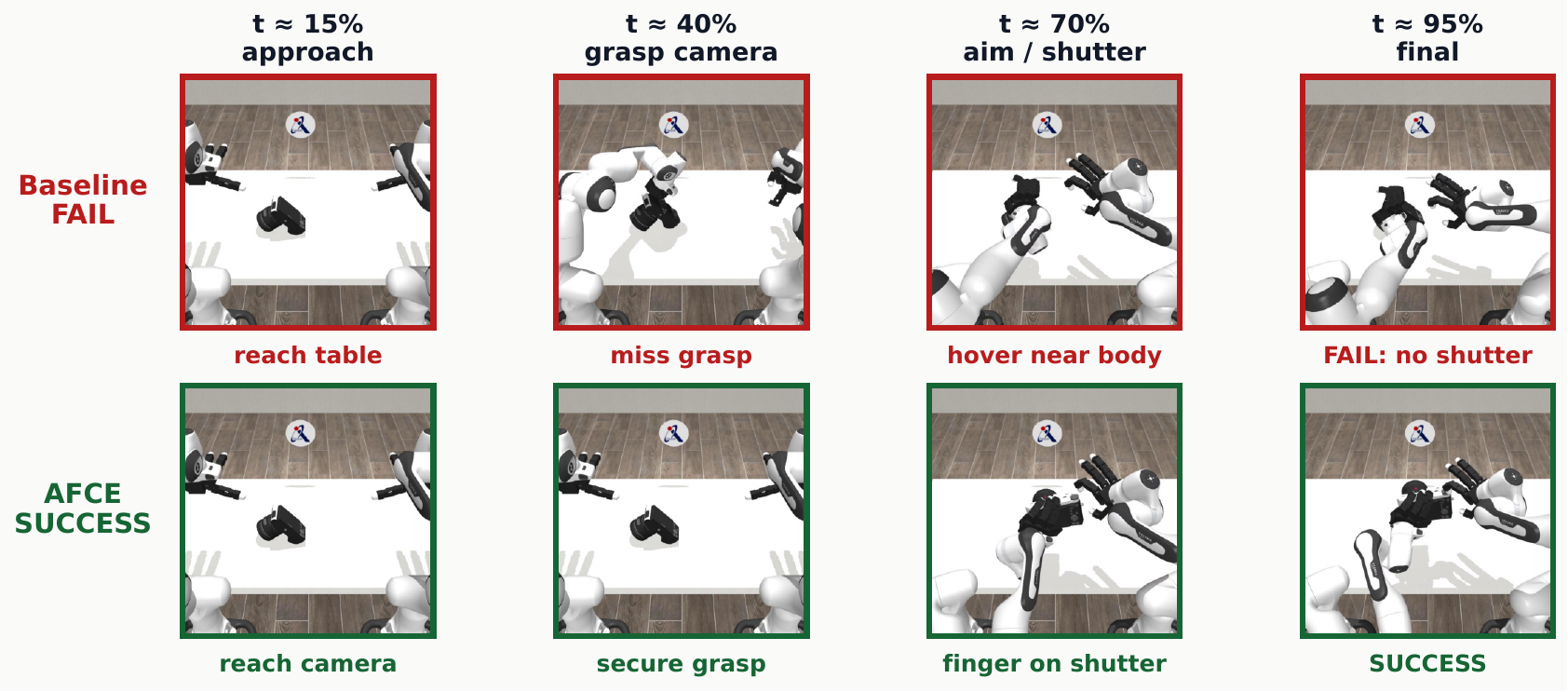}
  \caption{Illustrative AFCE and $\pi_{0.5}$ rollouts on Fold Glasses, Hanoi, Microwave, and Photograph (top to bottom). Each task is shown at approximately 15\%, 40\%, 70\%, and 95\% of the rollout. The selected AFCE rollouts complete their task-specific interaction, whereas the corresponding $\pi_{0.5}$ rollouts fail at contact, placement, or final actuation.}
  \label{fig:app-process-comparisons}
\end{figure*}

\paragraph{Qualitative task-progress comparisons.} Figure~\ref{fig:app-process-comparisons} compares illustrative AFCE and baseline rollouts at four relative progress points. In Fold Glasses, the baseline contacts the frame but leaves the glasses on the table, whereas AFCE folds and places them in the tray. In Hanoi, AFCE maintains a stable grasp and aligns the piece with the target peg, while the baseline ends with a tilted placement. AFCE also completes the ordered interactions in Microwave, progressing from the door to the controls, and in Photograph, progressing from camera grasp to shutter actuation. In the corresponding baseline rollouts, the microwave interaction remains incomplete and the shutter is not triggered. Together, these examples illustrate AFCE's ability to sustain task progress through contact, object transfer, and final actuation, complementing the task-level success rates in Table~\ref{tab:dexjoco-main}.

\section{Reproducibility Details}
\label{app:reproducibility}

\subsection{Implementation and Hyperparameters}

Table~\ref{tab:app-afce-implementation} lists the evaluated AFCE configuration. Stage I uses three 64-dimensional evidence branches and produces a $30\times256$ Effect. Its QueryFusion and temporal modules each have two layers, eight attention heads, a feed-forward width of 1024, and zero dropout. The state-conditioned action decoder has four layers with the same width, head count, and feed-forward width.

\begin{table}[t]
    \centering
    \caption{Hyperparameters of the evaluated AFCE configuration.}
    \label{tab:app-afce-implementation}
    \footnotesize
    \setlength{\tabcolsep}{4pt}
    \renewcommand{\arraystretch}{1.13}
    \begin{tabularx}{\linewidth}{@{}>{\raggedright\arraybackslash}p{0.28\linewidth}>{\raggedright\arraybackslash}X@{}}
        \toprule
        \textbf{Hyperparameter} & \textbf{Value} \\
        \midrule
        \multicolumn{2}{@{}l}{\textit{Stage I: Effect tokenizer}} \\
        Horizon / feature widths & $H=30$; branch width 64; Effect width 256 \\
        Fusion / temporal / decoder layers & $2/2/4$; 8 heads; FFN width 1024; dropout 0 \\
        Visual sampling & Stride 5 plus terminal frame; at most 6 dynamic and 2 reference regions, with 8 slots per region \\
        Visual descriptor / PCA & 136 dimensions / 64 components; frozen DINOv3; \texttt{legacy-none} mask \\
        Action reconstruction & Grouped Huber; position weight $w_p=2$ \\
        Auxiliary reconstruction & $\lambda_S(u)=\lambda_V(u)=0.05+0.15\min\{u/1000,1\}$ \\
        SO(3) loss weight & $\lambda_R=0.05$ \\
        Updates / global batch / seed & $30{,}000/32/42$ \\
        AdamW / gradient clip & $(\beta_1,\beta_2)=(0.9,0.99)$; weight decay 0.01; global norm 1.0 \\
        Learning rate & $10^{-4}$ through update 20,000; $3\times10^{-5}$ thereafter \\
        Effect normalization & Per-channel training-data statistics; standard-deviation floor $10^{-4}$ \\
        \midrule
        \multicolumn{2}{@{}l}{\textit{Stage II: Effect policy and action decoder}} \\
        Base / Effect shape & Official $\pi_{0.5}$ base; normalized Effect $30\times256$ \\
        Observation / instruction & One image time step; at most 250 language tokens; discrete state input enabled \\
        LoRA rank / alpha & PaliGemma 2B: $16/16$; 300M action expert: $32/32$ \\
        Flow-time sampling & $\tau=0.999\xi+0.001$, $\xi\sim\mathrm{Beta}(1.5,1)$ \\
        GT-action loss & Grouped Huber threshold 1; $\lambda_{\mathrm{GT}}=0.25628781345139295$ \\
        Finger-difference loss & Huber threshold 1; $\lambda_{\Delta f}=0.05$; per-joint RMS floor 0.1 \\
        Auxiliary schedule & $\tau\in[0.2,0.7]$; linear ramp over 10,000 updates \\
        Decoder learning rate & Policy learning rate $\times0.25$; decoder warm-up 0 updates \\
        Updates / global batch / seed & $60{,}000/32/42$ \\
        AdamW / gradient clip & $(\beta_1,\beta_2)=(0.9,0.95)$; $\epsilon=10^{-8}$; weight decay $10^{-10}$; global norm 1.0 \\
        Policy learning rate & Warm-up over 10,000 updates to $5\times10^{-5}$; constant thereafter \\
        Precision / EMA & Policy bfloat16; decoder and losses float32; no EMA \\
        Flow steps at inference & 10 \\
        \bottomrule
    \end{tabularx}
\end{table}

The Stage-I visual descriptors are projected with PCA fitted on tokenizer training episodes only. The \texttt{legacy-none} mask does not exclude robot regions. For Stage II, the original action input and output projections are reinitialized for the Effect width. Non-LoRA language-model weights are frozen. The Stage-I decoder is transferred to the policy implementation with separate single-arm and bimanual output interfaces; the corresponding decoder is trainable from update zero and is also used for execution.

\subsection{Training and Inference Procedure}

The two stages follow this procedure:
\begin{enumerate}
    \item \textbf{Train the Effect tokenizer.} Construct synchronized action, response, and sparse visual evidence for each demonstration window. Optimize the Stage-I objective in Eq.~\eqref{eq:app-effect-objective}. Use the tokenizer checkpoint at update 30,000 and report physical reconstruction on the 110 diagnostic trajectories drawn from its training demonstrations. These diagnostics are not held-out validation.

    \item \textbf{Cache fixed Effect targets.} Encode the demonstration windows using the selected tokenizer and store the resulting Effects in float16. Fit the channel-wise Effect normalization statistics using tokenizer training episodes only. Freeze the tokenizer, cached targets, and Effect, action, and state normalization statistics for Stage II.

    \item \textbf{Train the Effect policy and decoder.} Sample demonstration windows in proportion to their frame counts. For each batch, sample flow time and Gaussian noise, construct the noisy Effect, and compute flow matching. Decode the single-step clean Effect estimate and apply the ground-truth-action and finger-difference losses when the flow-time gate is active. Update the policy and the same embodiment decoder jointly using Eq.~\eqref{eq:app-afce-policy-objective}.

    \item \textbf{Execute the policy.} Starting from noise, perform ten flow steps to predict a normalized Effect. Reverse Effect normalization and pass it, together with the current robot state, to the jointly trained embodiment decoder. Execute the resulting native action chunk with receding-horizon replanning.
\end{enumerate}

For the two Stage-II action losses, the time weight is zero outside $\tau\in[0.2,0.7]$ and corrects for the probability mass of this interval under the specified flow-time sampling distribution. The GT-action coefficient was fixed after calibration on 16 batches: its value gives an approximately 0.1 weighted median gradient ratio relative to flow matching. Finger-difference scaling uses the batch RMS of demonstrated adjacent differences for each finger joint; gradients do not pass through that scale. Adjacent pairs are included only when both steps satisfy the Effect-valid mask.

Stage-II training uses two nodes with four GH200 GPUs each and a global batch size of 32. Sampling is frame proportional, with a committed data cursor, a PCG64 permutation for each epoch, and \texttt{drop\_last}. We evaluate the checkpoint obtained after 60,000 training updates. Training uses seed 42. Closed-loop results are evaluated with environment seeds 0, 1, and 2, using 50 episodes per task per seed and one trained checkpoint per method. The reported standard deviations measure variation across evaluation seeds, not across independent training runs. Task definitions, camera selection, episode counts, and evaluation metrics are specified in Sec.~\ref{app:evaluation-details}.



\end{document}